\documentclass[11pt]{article}
\usepackage{amsmath,amsthm,amsfonts}
\usepackage{graphicx}

\usepackage[dvips]{color}

\theoremstyle{definition}
\date{}
\newtheorem{Theorem}{\quad Theorem}[section]
\newtheorem{Definition}[Theorem]{\quad Definition}
\newtheorem{Corollary}[Theorem]{\quad Corollary}
\newtheorem{Proposition}[Theorem]{\quad Proposition}
\newtheorem{Lemma}[Theorem]{\quad Lemma}
\newtheorem{Example}[Theorem]{\quad Example}

\newtheorem{Algorithm}[Theorem]{\quad Algorithm}

\theoremstyle{definition}

\theoremstyle{remark}

\def\se{\vskip3pt plus1pt minus1pt\setbox0=\hbox to\hsize\bgroup\hss
        \vrule width.5pt
        \vbox\bgroup \hrule width \hsize height.5pt
        \vskip3pt\hbox to\hsize\bgroup\hss\vbox\bgroup\advance\hsize by-9pt
        \columnwidth\hsize\small}
\def\ee{\par\egroup\hss\egroup\vskip3pt\hrule width\hsize height.5pt\egroup
        \vrule width.5pt\hss\egroup
        \box0 \vskip3pt plus1pt minus1pt}
\def\latexe{\LaTeX\kern.15em 2${}_{\textstyle\varepsilon}$}
\title{}
\author{}

\begin{document}
\maketitle \markboth{} {}
\renewcommand{\sectionmark}[1]{}

\thispagestyle{empty}
\begin{center}
{\large{\textbf{Higher order hesitant fuzzy Choquet integral operator and its application to multiple criteria decision making
}}}
\\
\vspace{0.5cm} {\bf B. Farhadinia}\footnote{Corresponding author.} \quad  {\bf U. Aickelin}$^\dagger$ \quad {\bf H.A. Khorshidi}$^\dagger$
 \vspace*{0cm}
\\
\small{Dept. Math., Quchan University of Technology, Iran.}\\
\small{\verb"bfarhadinia@qiet.ac.ir"}
\\
\small{{$^\dagger$ Dept. Computing and Information Systems, University of Melbourne, Australia.}}
\\
\small{\verb"uwe.aickelin@unimelb.edu.au"}
\\
\small{\verb"hadi.khorshidi@unimelb.edu.au"}
\end{center}

\bigskip

\noindent {\bf Abstract.} 
{
Generally, the criteria involved in a decision making problem are interactive or inter-dependent, and therefore aggregating them by the use of traditional operators which are based on additive measures is not logical. This verifies that we have to implement fuzzy measures for modelling the interaction phenomena among the criteria.
{On the other hand, based on the recent extension of hesitant fuzzy set, called higher order hesitant fuzzy
set (HOHFS) which allows the
membership of a given element to be defined in forms of several possible generalized types
of fuzzy set, we encourage to} 
propose the higher order hesitant fuzzy (HOHF) Choquet integral operator. {This concept} not only considers the importance of the higher order hesitant fuzzy arguments, but also it can reflect the correlations among those arguments. Then, 
a detailed discussion on the aggregation properties of the HOHF Choquet integral operator will be presented.
To enhance the application of HOHF Choquet integral operator in decision making, we first assess
the appropriate energy policy for the socio-economic development. Then, the efficiency of the proposed HOHF Choquet integral operator-based technique over a number of exiting techniques is further verified by employing another  
decision making problem associated with the technique of TODIM (an acronym in Portuguese of Interactive and Multicriteria Decision Making). }

\bigskip


%
\section{Introduction}

{The theory of extending and generalizing fuzzy sets is very prospective tool in different research domains.
One of interesting extensions is the concept of hesitant fuzzy set (HFS) which was proposed by Torra and Narukawa \cite{7@2}. HFS is  
quite suit for the case where the decision maker has a set of possible values,
rather than a margin of error (like that in intuitionistic fuzzy sets \cite{atan}) or some possibility
distribution on the possible values (like that is considered in defining type-2 fuzzy sets \cite{mizu}).
By the way, HFS \cite{7@2} and its extension which is called the generalized HFS (G-HFS) \cite{qian}, have their own drawbacks because of the reason that the membership degrees of an element to a given set can be expressed only by crisp numbers or intuitionistic fuzzy sets. 
The evidence to this fact is that in} 
a real and practical decision making problem {where} the information provided by a group of experts may not be described only by {hesitant fuzzy sets (HFSs) or} just by one kind of HFS extensions. 
A more specific situation occurs when a number of experts are intended to return their evaluations in the form of hesitant multiplicative sets, {and others are} interested to describe their evaluation by the help of hesitant fuzzy linguistic term sets. Therefore, it is difficult for the decision maker to
provide exact hesitant multiplicative set or hesitant fuzzy linguistic term set for the membership degrees. Such a difficulty is avoided by considering the higher order HFS (HOHFS) \cite{comp30} in order to describe the membership degrees. {The concept of HOHFS} not only encompasses fuzzy sets, intuitionistic fuzzy sets, type 2 fuzzy sets and HFSs, but also it extends the concept of generalized hesitant fuzzy {set (G-HFS)} \cite{qian}.


{It does worth to} say that the topic of developing information aggregations in fuzzy multiple criteria  decision
making (MCDM) has been an interesting research topic {\cite{farhadisco,farhadixia,far-book,x1,x2}} so far. As it is known, most of the existing fuzzy aggregation operators only consider those situations in which all the
elements are independent.
Moreover, in many real-world situations, we observe that
the elements  are usually interactive or interdependent, that is, they are correlative. 

The Choquet integral \cite{9} with respect to fuzzy measures \cite{43} is an aggregation function that has been employed successfully in MCDM  problems where {it determines the} relative importance of decision criteria as well as their interactions \cite{32}. 
There exist a growing number of studies on the Choquet integral, and it has been investigated by many researchers.
Yager \cite{[43]} proposed the induced Choquet
ordered averaging operator, and moreover, he introduced the concept of 
Choquet aggregation based on the order induced aggregation.
By extending the concept of  Choquet integral to that for fuzzy context,  Meyer and Roubens
\cite{[44]} applied this concept to multiple criteria decision making problems.
Yu et al. \cite{[45]} defined a hesitant fuzzy aggregation
operator by taking the concept of  Choquet integral into account, and they implemented it in modelling a hesitant-based multiple criteria decision making problem.
Tan and Chen \cite{[46]} and Tan \cite{[47]} proposed respectively the intuitionistic and the interval-valued intuitionistic fuzzy Choquet integral
operators. 
Bustince et al. \cite{[48]} investigated a multiple criteria decision making technique by considering the concept of interval-valued Choquet
integral. 
Wang et al. \cite{[50]} investigated multiple criteria group decision making problems in which the Choquet integral
aggregation operators together with interval two-tuple linguistic
information play the main role.
{Some more studies have been given by Abdullah et al. \cite{ex2}, Candeloro et al. \cite{ex1}, Pasi et al. \cite{ex3}, etc.}

By the way, inspired by the concept of Choquet integral, we will develop here a higher order hesitant fuzzy Choquet
integral operator for aggregating higher order hesitant fuzzy (HOHF) information
in MCDM, and investigate different
properties of this operator.
{
As will be shown in Section 3, such a proposed concept not only is well applicable to the case where 
the decision makers have a hesitation among several possible memberships
with uncertainties in a general form (i.e., HOHFSs), but also it models the interaction phenomena among the criteria (i.e., the Choquet
integral property).}

The present paper is organized as follows: The concept of  higher order HFS (HOHFS)  as an extension of HFSs is reviewed in Section 2, and then the higher order hesitant fuzzy (HOHF) Choquet integral operator is presented together with a detailed discussion on the aggregation properties of the HOHF Choquet integral operator. In Section 3, we apply the proposed HOHF Choquet integral operator to MCDM problems involving the higher order hesitant fuzzy information, and then a comparative analysis is then given to demonstrate the effectiveness of the proposed operator. Finally, conclusion is drown in Section 4.

\section{The HOHF Choquet integral operator}
Among the several extensions of the theory of HFSs, we deal in brief with the higher order hesitant fuzzy set (HOHFS) which was first introduced by Farhadinia \cite{comp30}.
%
\begin{Definition}\label{Def2.1}\cite{comp30}
Let $X$ be the universe of discourse. A generalized type of fuzzy set (G-Type FS) on $X$ is defined as
\begin{eqnarray}\label{gtfs}
\widetilde{A}=\{\langle x,\widetilde{A}(x)\rangle:x\in X\},
\end{eqnarray}
where
\begin{eqnarray*}\label{gtfsc}
\widetilde{A}:X\rightarrow \psi([0,1]).
\end{eqnarray*}
\end{Definition}
Here, $\psi([0,1])$ denotes {a family of crisp or fuzzy sets} that can be defined with in the universal set $[0,1]$.\\
We mention here some special cases of G-Type FS which are encountered by Klir and Yuan in \cite{klir}: 
\begin{itemize}
  \item if $\psi([0,1])=[0,1]$, then the G-Type FS $\widetilde{A}$ reduces to an ordinary fuzzy set;
  \item if $\psi([0,1])=\varepsilon([0,1])$ denoting the set of all closed intervals, then the G-Type FS $\widetilde{A}$ reduces to an interval-valued fuzzy set;
  \item if $\psi([0,1])={\cal F}([0,1])$ denoting the set of all ordinary fuzzy sets, then the G-Type FS $\widetilde{A}$ reduces to a type II fuzzy set;
  \item if $\psi([0,1])=L$ denoting a partially ordered Lattice, then the G-Type FS $\widetilde{A}$ reduces to a lattice fuzzy set.
\end{itemize}
\begin{Definition}\label{Def2.2}
\cite{7@2} Let $X$ be the universe of discourse. A hesitant fuzzy set (HFS) on $X$ is symbolized by
$$H=\{\langle x,h(x)\rangle:x\in X\},$$
where $h(x)$, referred to as the hesitant fuzzy element (HFE), is a set of some values in $[0, 1]$ denoting the possible membership degree of the element $x\in X$ to the set $H$.
\end{Definition}
As can be seen from Definition \ref{Def2.2}, HFS expresses the membership degrees of an element to a given set only by several real numbers between 0 and 1, while in many real-world situations assigning exact values to the membership degrees does not describe properly the imprecise or uncertain decision information.
Thus, it seems to be difficult for the decision makers to rely on HFSs for expressing uncertainty of an element.
\\
To overcome the difficulty associated with expressing uncertainty of an element to a given set, the {concept of HOHFS
allows} the membership degrees of an element {to be} expressed by several possible G-Type FSs.
\begin{Definition}\label{Def2.3}\cite{comp30}
Let $X$ be the universe of discourse. A higher order hesitant fuzzy set (HOHFS) on $X$ is defined
in terms of a {function when it is applied to $X$, it then returns} a set of G-Type FSs. A HOHFS is denoted by
\begin{eqnarray}\label{hohfs}
\widetilde{H}=\{\langle x,\widetilde{h}(x)\rangle:x\in X\},
\end{eqnarray}
where $\widetilde{h}(x)$, referred to as the higher order hesitant fuzzy element (HOHFE), is a set of some G-Type FSs denoting the possible membership degree of the element $x\in X$ to the set $\widetilde{H}$. In this regards, the HOHFS $\widetilde{H}$ is also represented as
\begin{eqnarray*}\label{-}
\widetilde{H}=\{\langle x,\{\widetilde{h_{\widetilde{h}}}^{(1)}(x),...,\widetilde{h_{\widetilde{h}}}^{(|\widetilde{h}(x)|)}(x)\}\rangle:x\in X\},
\end{eqnarray*}
where all $\widetilde{h_{\widetilde{h}}}^{(1)}(x),...,\widetilde{h_{\widetilde{h}}}^{(|\widetilde{h}(x)|)}(x)$ are G-Type FSs on $X$, and $|\widetilde{h}(x)|$ is the number of G-type FSs in $\widetilde{h}(x)$.
\end{Definition}
The noteworthy feature about Definition \ref{Def2.3} is that it encompasses all the above-mentioned generalizations of ordinary fuzzy sets which {have} been discussed in detail by Farhadinia \cite{comp30}.

Having introduced HOHFEs of a HOHFS $\widetilde{H}=\{\langle x,\widetilde{{h}}(x)\rangle:x\in X\}$, we now turn our attention to the interpretation of HOHFS $\widetilde{H}$ as the union of all HOHFEs
i.e.,
{
\begin{eqnarray}\label{hrep}
\widetilde{H}=\bigcup_{\widetilde{{h}}(x)\in \widetilde{H}}\{\widetilde{{h}}(x)\}
\end{eqnarray}
which is fundamental in} the study of HOHFS aggregation operators within the next parts of the paper.

{Hereafter, to simplify the notation, we use only $\widetilde{{h}}$ instead of $\widetilde{{h}}(x)$ in the following descriptions, that is, we assume that
\begin{eqnarray*}
\widetilde{H}=\bigcup_{\widetilde{{h}}\in \widetilde{H}}\{\widetilde{{h}}\}.
\end{eqnarray*}
}

\begin{Definition}\label{Def3.1}
Let $\widetilde{H}_1=\bigcup_{\widetilde{{h}}_1\in \widetilde{H}_1}\{\widetilde{{h}_1}\}$ and $\widetilde{H}_2=\bigcup_{\widetilde{{h}}_2\in \widetilde{H}_2}\{\widetilde{{h}_2}\}$ be two HOHFSs. We give the definition of operation $\odot$ on HOHFSs based on its counterpart on HOHFEs as follows:
\begin{eqnarray}\label{op}
\widetilde{H}_1\odot\widetilde{H}_2=\bigcup_{\widetilde{h_1}\in \widetilde{H_1},\widetilde{h_2}\in \widetilde{H_2}}\{\widetilde{{h_1}}\odot\widetilde{{h_2}}\}
=\bigcup_{\widetilde{h_1}\in \widetilde{H_1},\widetilde{h_2}\in \widetilde{H_2}}\{\bigcup_{\widetilde{h_{h1}}\in \widetilde{h_1},\widetilde{h_{h2}}\in \widetilde{h_2}}
\{
\widetilde{{h_{h1}}}\odot\widetilde{{h_{h2}}}\}~\},
\end{eqnarray}
where the last operation $\odot$ stands for the operation on G-type FSs.
\end{Definition}

Note that, if there is no confusion, 
{all the operations performing over HOHFSs, HOHFEs  and G-type FSs}  are denoted by the same symbol. {Furthermore, that property holds for G-type FSs is denoted by $\widetilde{(p_G)}$, and that is defined on HOHFEs is symbolized by $\widetilde{(p)}$.}

\begin{Theorem}\label{Th3.1}
Let $\widetilde{h}_1=\bigcup_{\widetilde{{h}}_{h1}\in \widetilde{h}_1}\{\widetilde{{h}_{h1}}\}$ and $\widetilde{h}_2=\bigcup_{\widetilde{{h}}_{h2}\in \widetilde{h}_2}\{\widetilde{{h}_{h2}}\}$ be two HOHFEs. If the operation $\odot$ on G-type FSs has the property $\widetilde{(p_G)}$, then the operation $\odot$ on HOHFEs has the {counterpart} property, denoted by $\widetilde{(p)}$.
\end{Theorem}
\textbf{Proof.} The proof is obvious from Definition \ref{Def3.1} Suppose that, for example, the operation $\odot$ on G-type FSs is commutative, that is, for any G-type FSs $\widetilde{h_{h1}}\in \widetilde{h_1}$ and $\widetilde{h_{h2}}\in \widetilde{h_2}$, we have
\begin{eqnarray}\nonumber
\widetilde{{h}_{h1}}\odot\widetilde{{h}_{h2}}=\widetilde{{h}_{h2}}\odot\widetilde{{h}_{h1}}.\quad\quad(\textrm{Property} ~\widetilde{(p_G)})
\end{eqnarray}
Then, one can see from Definition  \ref{Def3.1}  that
\begin{eqnarray}\nonumber
\widetilde{h}_1\odot\widetilde{h}_2=\bigcup_{\widetilde{h_{h1}}\in \widetilde{h_1},\widetilde{h_{h2}}\in \widetilde{h_2}}\{\widetilde{{h_{h1}}}\odot\widetilde{{h_{h2}}}\}=
\bigcup_{\widetilde{h_{h1}}\in \widetilde{h_1},\widetilde{h_{h2}}\in \widetilde{h_2}}\{\widetilde{{h_{h2}}}\odot\widetilde{{h_{h1}}}\}=\widetilde{h}_2\odot\widetilde{h}_1.\quad\quad(\textrm{Property} ~\widetilde{(p)})
\end{eqnarray}
This means that the operation $\odot$ on HOHFEs is accordingly commutative.\\
By similar reasoning, one can easily prove that the HOHFE operator inherits all the properties of the G-type FS operator. $\Box$

As a corollary of Theorem  \ref{Th3.1} , it can be observed that HOHFS operators inherit subsequently all operational properties of G-type FS operators.

\begin{Corollary}\label{Co3.1}
Let $\widetilde{H}_1=\bigcup_{\widetilde{{h}}_1\in \widetilde{H}_1}\{\widetilde{{h}_1}\}$ and $\widetilde{H}_2=\bigcup_{\widetilde{{h}}_2\in \widetilde{H}_2}\{\widetilde{{h}_2}\}$ be two HOHFSs. If the operation $\odot$ on HOHFEs has the property $\widetilde{(p)}$, then the operation $\odot$ on HOHFSs has the mentioned property, denoted by $\widetilde{(P)}$.
\end{Corollary}

Ranking fuzzy information play a key role in fuzzy decision-making procedure. This fact is very encouraging for introducing a score function of HOHFSs. To compare the HOHFSs, we define the following comparison laws:
\begin{Definition}\label{Def3.2}
Let $X=\{x_1,x_2...,x_n\}$. For HOHFS $\widetilde{H}=\{\langle x,\widetilde{{h}}(x)\rangle:x\in X\}=
\bigcup_{\widetilde{h}\in \widetilde{H}}\{\bigcup_{\widetilde{h_{h}}\in \widetilde{h}}\{
\widetilde{{h_{h}}}\}\}$, the score function ${\cal S}(.)$ is defined by
\begin{eqnarray}\label{scor}
{\cal S}(\widetilde{H})=\frac{1}{n}\sum_{i=1}^{n}S_{\widetilde{h}}(\widetilde{h}(x_i))=
\frac{1}{n}\sum_{i=1}^{n}(~\frac{1}{|\widetilde{h}(x_i)|}\sum_{\widetilde{h_{h}}\in \widetilde{h}(x_i)}S_{\widetilde{{h_{h}}}}(\widetilde{{h_{h}}}(x_i))~),
\end{eqnarray}
where $S_{\widetilde{h}}(.)$ is a score function of HOHFEs, $|\widetilde{h}(x_i)|$ is the number of G-type FSs in $\widetilde{h}(x_i)$ and $S_{\widetilde{{h_{h}}}}(.)$ is a score function of G-type FSs.
For two HOHFSs $\widetilde{H_1}$ and $\widetilde{H_2}$, if ${\cal S}(\widetilde{H}_1)<{\cal S}(\widetilde{H}_2)$ then $\widetilde{H_1}\prec\widetilde{H_2}$; if ${\cal S}(\widetilde{H}_1)={\cal S}(\widetilde{H}_2)$ then $\widetilde{H_1}\approx\widetilde{H_2}$.
\end{Definition}

Similar comparison rules can be considered for HOHFEs and G-type FSs such that
\begin{itemize}
  \item for two HOHFEs $\widetilde{h_1}$ and $\widetilde{h_2}$ if $S_{\widetilde{h}}(\widetilde{h}_1)<S_{\widetilde{h}}(\widetilde{h}_2)$ then $\widetilde{h_1}\prec\widetilde{h_2}$; if $S_{\widetilde{h}}(\widetilde{h}_1)=S_{\widetilde{h}}(\widetilde{h}_2)$ then $\widetilde{h_1}\approx\widetilde{h_2}$;
\item for two G-type FSs $\widetilde{h_{h}}^{(1)}$ and $\widetilde{h_{h}}^{(2)}$ if $S_{\widetilde{h_h}}(\widetilde{h}_{h}^{(1)})<S_{\widetilde{h_h}}(\widetilde{h}_{h}^{(2)})$ then $\widetilde{h_{h}}^{(1)}\prec\widetilde{h_{h}}^{(2)}$; if $S_{\widetilde{h_h}}(\widetilde{h}_{h}^{(1)})=S_{\widetilde{h_h}}(\widetilde{h}_{h}^{(2)})$ then $\widetilde{h_{h}}^{(1)}\approx\widetilde{h_{h}}^{(2)}$.
\end{itemize}

In most of the real world applications, we usually face to the elements in the universe of discourse which may have a
different importance. This impulses us to consider the weight of each element $x_i\in X$. Assume that the weight of $x_i\in X$ is $\mu(\{x_i\})$, $(i=1,...,n)$ where $\mu$ is a \textit{fuzzy measure}.
\begin{Definition}\label{Def3.3}
\cite{43} A fuzzy measure on $X$ is a set function $\mu$ from the power set of $X$ into $[0,1]$, satisfying the following conditions:
\begin{eqnarray*}\nonumber
&&(1)~ \mu(\emptyset)=0,~\mu(X)=1;\\
&&(2)~ \textrm{For}~\textrm{any}~A,B\subseteq X,~  \textrm{if}~A\subseteq B,~\textrm{then}~\mu(A)\leq \mu(B);\\
&&(3)~  \textrm{For}~\textrm{any}~A,B\subseteq X, ~\mu(A\cup B)=\mu(A)+\mu(B)-\rho\mu(A)\mu(B),~\textrm{where} ~A\cap B=\emptyset~ \textrm{and}~\rho\in(-1,\infty).
\end{eqnarray*}
\end{Definition}
A fuzzy measure $\mu$ on $X$ is said to be
\begin{itemize}
  \item \textit{Additive} if $\mu(A\cup B)=\mu(A)+\mu(B)$ for all $A,B\subseteq X$ where $A\cap B=\emptyset$;
  \item \textit{Subadditive} if $\mu(A\cup B)\leq\mu(A)+\mu(B)$ for all $A,B\subseteq X$ where $A\cap B=\emptyset$;
  \item \textit{Superadditive} if $\mu(A\cup B)\geq\mu(A)+\mu(B)$ for all $A,B\subseteq X$ where $A\cap B=\emptyset$.
\end{itemize}

Sometimes, {in MCDM}, we need an aggregation function {to determine the} relative importance of decision criteria {considering} their interactions. Such a need could be satisfied by implementing the Choquet integral \cite{9} which is defined on the basis of fuzzy measures.
\begin{Definition}\label{Def3.4}
\cite{9} Let $f$ be a positive real-valued
function on $X$, and $\mu$ be a fuzzy measure on $X$. The Choquet
integral of $f$ with respect to $\mu$ is defined by
\begin{eqnarray}\label{cmudis}
C_\mu(f)=\sum_{i=1}^{n}(\mu(A_{\sigma(i)})-\mu(A_{\sigma(i-1)})f_{\sigma(i)}
\end{eqnarray}
where $(\sigma(1),\sigma(2), . . . ,\sigma(n))$ is a permutation of $(1, 2, . . . ,n)$ such that $f_{\sigma(1)}\geq f_{\sigma(2)}\geq...\geq f_{\sigma(n)}$, $A_{\sigma(k)} =\{x_{\sigma(1)},x_{\sigma(2)},...,x_{\sigma(k)}\}$ for
$k\geq 1$, and $A_{\sigma(0)} =\emptyset$.
\end{Definition}
The Choquet integral $C_\mu(.)$ has some considerable aggregation properties, such as idempotence, compensativeness and comonotonic
additivity \cite{32}. Moreover, it generalizes the weighted arithmetic mean (WAM) and the ordered weighted averaging (OWA) \cite{yagerid}. In view of the Choquet integral formulation (\ref{cmudis}), we here introduce the HOHF Choquet integral:
\begin{Definition}\label{Def3.5}
Let $\mu$ be a fuzzy measure on $X$, and $\widetilde{h}(x_i),~(i=1,...,n)$ be a collection of
HOHFEs on $X$. The HOHF Choquet integral of $\widetilde{h}(x_i),~(i=1,...,n)$ with
respect to $\mu$ is defined by
\begin{eqnarray}\nonumber
&&HOHFEC_\mu(\widetilde{h}(x_1),...,\widetilde{h}(x_n))=\oplus_{i=1}^{n}[(\mu(A_{\sigma(i)})-\mu(A_{\sigma(i-1)})\widetilde{h}(x_{\sigma(i)})]\hspace{5cm}\\
&&\hspace{1cm}=\bigcup_{\widetilde{h_h}(x_{\sigma(1)})\in\widetilde{h}(x_{\sigma(1)}),...,\widetilde{h_h}(x_{\sigma(n)})\in\widetilde{h}(x_{\sigma(n)})}
\{\oplus_{i=1}^{n}[(\mu(A_{\sigma(i)})-\mu(A_{\sigma(i-1)})\widetilde{h_h}(x_{\sigma(i)})]\},
\label{hocmu1}
\end{eqnarray}
where $(\sigma(1),\sigma(2), . . . ,\sigma(n))$ is a permutation of $(1, 2, . . . ,n)$ such that
$\widetilde{h}(x_{\sigma(1)})\succ \widetilde{h}(x_{\sigma(2)})\succ...\succ \widetilde{h}(x_{\sigma(n)})$,
$A_{\sigma(k)} =\{x_{\sigma(1)},x_{\sigma(2)},...,x_{\sigma(k)}\}$ for
$k\geq 1$, and $A_{\sigma(0)} =\emptyset$.
\end{Definition}
\begin{Proposition}\label{Pr3.1}
If for any G-type FSs $\widetilde{h_h}^{(1)}$ and $\widetilde{h_h}^{(2)}$ of HOHFE $\widetilde{h}$, the addition $\widetilde{h_h}^{(1)}\oplus\widetilde{h_h}^{(2)}$ and the scalar multiplication $\lambda\widetilde{h_h}^{(1)}$ where $\lambda>0$ are also G-type FSs, then the HOHF Choquet integral of $\widetilde{h}(x_i),~(i=1,...,n)$ with
respect to $\mu$ introduced in Definition \ref{Def3.5} is a HOHFE.
\end{Proposition}
\textbf{Proof.} The proof is obvious by putting together the definition of the HOHF Choquet integral of
$\widetilde{h}(x_i),~(i=1,...,n)$ with respect to $\mu$ introduced in Definition \ref{Def3.5} and Theorem \ref{Th3.1}. $\Box$
\begin{Example}\label{Ex3.1}
Let $X = \{x_1,x_2\}$, and $\widetilde{h}(x_i),~(i=1,2)$ be a collection of
HOHFEs on $X$ whose G-type FSs are in the form of IFSs given by
\begin{eqnarray*}\nonumber
\widetilde{h}(x_1)=\{\widetilde{h_h}^{(1)}(x_1)=\langle0.2,0.5\rangle , \widetilde{h_h}^{(2)}(x_1)=\langle0.3,0.4\rangle\}, \quad \quad
\widetilde{h}(x_2)=\{\widetilde{h_h}^{(1)}(x_2)=\langle0.3,0.5\rangle \},
\end{eqnarray*}
and a fuzzy measure $\mu: X\rightarrow[0,1]$ is given by
\begin{eqnarray*}\nonumber
\mu(\emptyset)=0,\quad\mu(\{x_1\})=0.2,\quad\mu(\{x_2\})=0.3,\quad\mu(\{x_1,x_2\})=0.6.
\end{eqnarray*}
If we take the G-type FSs score function as the hesitancy degree, i.e.,
\begin{eqnarray*}\nonumber
S_{\widetilde{h_h}}(\widetilde{h_h})=S_{\widetilde{h_h}}(\langle\mu_{\widetilde{{h_{h}}}},\nu_{\widetilde{{h_{h}}}}\rangle)=
1-\mu_{\widetilde{{h_{h}}}}-\nu_{\widetilde{{h_{h}}}},
\end{eqnarray*}
then the HOHFE scores $S_{\widetilde{h}}(\widetilde{h}(x_1))=0.3$ and $S_{\widetilde{h}}(\widetilde{h}(x_2))=0.1$ indicate that $\widetilde{h}(x_1)\succ \widetilde{h}(x_2)$ and therefore ${\sigma(1)}=1,{\sigma(2)}=2$.
\\
Now from (\ref{hocmu1}) the HOHF Choquet integral of $\widetilde{h}(x_i),~(i=1,2)$ with
respect to $\mu$ is calculated as follows:
\begin{eqnarray*}\nonumber
&&HOHFEC_\mu(\widetilde{h}(x_1),\widetilde{h}(x_2))
=\bigcup_{\widetilde{h_h}(x_1)\in\widetilde{h}(x_1),\widetilde{h_h}(x_2)\in\widetilde{h}(x_2)}
\{\oplus_{i=1}^{2}[(\mu(A_{\sigma(i)})-\mu(A_{\sigma(i-1)})\widetilde{h_h}(x_{\sigma(i)})]\}\\
&&=\{[(\mu(A_1)-\mu(A_0)]\widetilde{h_h}^{(1)}(x_1)\oplus[(\mu(A_2)-\mu(A_1)]\widetilde{h_h}^{(1)}(x_2),\\
&&\hspace{4cm}[(\mu(A_1)-\mu(A_0)]\widetilde{h_h}^{(2)}(x_1)\oplus[(\mu(A_2)-\mu(A_1)]\widetilde{h_h}^{(1)}(x_2)\}\\
&&=\{[(\mu(\{x_1\})-\mu(\emptyset)]\widetilde{h_h}^{(1)}(x_1)\oplus[(\mu(\{x_2\})-\mu(\{x_1\})]\widetilde{h_h}^{(1)}(x_2),\\
&&\hspace{4cm}[(\mu(\{x_1\})-\mu(\emptyset)]\widetilde{h_h}^{(2)}(x_1)\oplus[(\mu(\{x_2\})-\mu(\{x_1\})]\widetilde{h_h}^{(1)}(x_2)\}\\
&&=\{[0.2]\langle0.2,0.5\rangle\oplus[0.6-0.2]\langle0.3,0.5\rangle,\\
&&\hspace{4cm}[0.2]\langle0.3,0.4\rangle\oplus[0.6-0.2]\langle0.3,0.5\rangle\}.
\end{eqnarray*}
By the use of {the following relations (see e.g. \cite{ifsop}) 
\begin{eqnarray*}
&&\widetilde{{h_{h1}}}\oplus\widetilde{{h_{h2}}}=\langle\mu_{\widetilde{{h_{h1}}}},\nu_{\widetilde{{h_{h1}}}}\rangle
\oplus\langle\mu_{\widetilde{{h_{h2}}}},\nu_{\widetilde{{h_{h2}}}}\rangle
=\langle\mu_{\widetilde{{h_{h1}}}}+\mu_{\widetilde{{h_{h2}}}}-\mu_{\widetilde{{h_{h1}}}}.\mu_{\widetilde{{h_{h2}}}},~~ \nu_{\widetilde{{h_{h1}}}}.\nu_{\widetilde{{h_{h2}}}}\rangle;\label{ifsops1}\\
&&\widetilde{{h_{h1}}}^\lambda=\langle(\mu_{\widetilde{{h_{h1}}}})^\lambda,1-(1-\nu_{\widetilde{{h_{h1}}}})^\lambda\rangle,\quad \lambda>0,\label{ifsops2}
\end{eqnarray*}
in which $\mu_{\widetilde{{h_{h2}}}}$ and $\nu_{\widetilde{{h_{h2}}}}$ stand respectively for the membership and the non-membership functions, we get}
\begin{eqnarray*}\nonumber
&&\widetilde{h}_{_{HOHFEC}}:=HOHFEC_\mu(\widetilde{h}(x_1),\widetilde{h}(x_2))
=\{\langle0.1708,0.6598\rangle,\langle0.1927,0.6310\rangle\}.
\end{eqnarray*}
\end{Example}

\begin{Proposition}\label{Pr3.2}
(\textit{Idempotency}). If for any G-type FS $\widetilde{h_h}$ of HOHFE $\widetilde{h}$, it holds $\lambda_1\widetilde{h_h}\oplus\lambda_2\widetilde{h_h}=(\lambda_1+\lambda_2)\widetilde{h_h}$ where $\lambda_1,\lambda_2>0$, and if all
HOHFEs $\widetilde{h}(x_i),~(i=1,...,n)$ are equal, that is, for all $i$, $\widetilde{h}(x_i)=\overline{\widetilde{h}}$,
then the HOHF Choquet integral of $\widetilde{h}(x_i),~(i=1,...,n)$ with
respect to $\mu$ introduced in Definition \ref{Def3.5} is equal to the HOHFE $\overline{\widetilde{h}}$.
\end{Proposition}
\textbf{Proof.} Suppose that for all $i,~(i=1,...,n)$, we have $\widetilde{h}(x_i)=\overline{\widetilde{h}}$. From definition of the HOHF Choquet integral of $\widetilde{h}(x_i),~(i=1,...,n)$ with
respect to $\mu$ introduced in Definition  \ref{Def3.5}, we have
\begin{eqnarray*}\nonumber
HOHFEC_\mu(\widetilde{h}(x_1),...,\widetilde{h}(x_n))&=&\oplus_{i=1}^{n}[(\mu(A_{\sigma(i)})-\mu(A_{\sigma(i-1)})\widetilde{h}(x_{\sigma(i)})]\\ \nonumber
&&=[(\mu(A_{\sigma(1)})-\mu(A_{\sigma(0)})]\overline{\widetilde{h}}\oplus...\oplus[(\mu(A_{\sigma(n)})-\mu(A_{\sigma(n-1)})]\overline{\widetilde{h}}.
\end{eqnarray*}
If for any G-type FS $\widetilde{h_h}$ of HOHFE $\widetilde{h}$, it holds $\lambda_1\widetilde{h_h}\oplus\lambda_2\widetilde{h_h}=(\lambda_1+\lambda_2)\widetilde{h_h}$ where $\lambda_1,\lambda_2>0$, then by Theorem \ref{Th3.1} and the fact that $\mu(A_{\sigma(0)}=\emptyset)=0,\mu(A_{\sigma(n)}=X)=1$, one gets
\begin{eqnarray*}\nonumber
HOHFEC_\mu(\widetilde{h}(x_1),...,\widetilde{h}(x_n))
&=([(\mu(A_{\sigma(1)})-\mu(A_{\sigma(0)})]+...+[(\mu(A_{\sigma(n)})-\mu(A_{\sigma(n-1)})])\overline{\widetilde{h}}=\overline{\widetilde{h}}.~\Box
\end{eqnarray*}
\begin{Lemma}\label{Le3.1}
Suppose that the score functions $S_{\widetilde{h}}$ on HOHFEs and $S_{\widetilde{{h_{h}}}}$ on G-type FSs are those introduced in Definition \ref{Def3.2}, that is,
\begin{eqnarray}\label{scors}
S_{\widetilde{h}}(\widetilde{h})=\frac{1}{|\widetilde{h}|}\sum_{\widetilde{h_{h}}\in \widetilde{h}}S_{\widetilde{{h_{h}}}}(\widetilde{{h_{h}}}).
\end{eqnarray}
If for each HOHFE $\widetilde{h}=\bigcup_{\widetilde{h_{h}}\in \widetilde{h}}\{\widetilde{h_{h}}\}$, $S_{\widetilde{{h_{h}}}}$ satisfies $S_{\widetilde{{h_{h}}}}(\lambda\widetilde{{h_{h}}})=\lambda S_{\widetilde{{h_{h}}}}(\widetilde{{h_{h}}})$, for any $\lambda>0$. Then, $S_{\widetilde{h}}$ also satisfies the latter property, i.e.,
\begin{eqnarray}\label{hamgen}
S_{ \widetilde{h}}(\lambda\widetilde{h})=\lambda S_{ \widetilde{h}}(\widetilde{h}),\quad \forall\lambda>0.
\end{eqnarray}
\end{Lemma}
\textbf{Proof.} From definition of $S_{\widetilde{h}}$ given by (\ref{scors}) and the mentioned property of $S_{\widetilde{{h_{h}}}}$, one gets for any $\lambda>0$
\begin{eqnarray*}
S_{\widetilde{h}}(\lambda\widetilde{h})=\frac{1}{|\lambda\widetilde{h}|}\sum_{\widetilde{h_{h}}\in \widetilde{h}}S_{\widetilde{{h_{h}}}}(\lambda\widetilde{{h_{h}}})=\frac{1}{|\widetilde{h}|}\sum_{\widetilde{h_{h}}\in \widetilde{h}}S_{\widetilde{{h_{h}}}}(\lambda\widetilde{{h_{h}}})=\frac{1}{|\widetilde{h}|}\sum_{\widetilde{h_{h}}\in \widetilde{h}}\lambda S_{\widetilde{{h_{h}}}}(\widetilde{{h_{h}}})=\lambda S_{\widetilde{h}}(\widetilde{h}).~\Box
\end{eqnarray*}
Hereafter, the HOHFE score function $S_{\widetilde{h}}$ is said to be \textit{$\lambda$-invariance} if it satisfies the property (\ref{hamgen}).
\begin{Lemma}\label{Le3.2}
Assume that the HOHFE score function $S_{\widetilde{h}}$ is $\lambda$-invariance. For any two HOHFEs $\widetilde{h}_1$ and $\widetilde{h}_2$, if $\widetilde{h}_1\succ\widetilde{h}_2$, then $\lambda\widetilde{h}_1\succ\lambda\widetilde{h}_2$, for any $\lambda>0$.
\end{Lemma}
\textbf{Proof.} As follows from Definition \ref{Def3.2}, we find that $\widetilde{h}_1\succ\widetilde{h}_2$ whenever $S_{\widetilde{h}}(\widetilde{h}_1)>S_{\widetilde{h}}(\widetilde{h}_2)$. On the other hand, with the $\lambda$-invariance property of the HOHFE score function $S_{\widetilde{h}}$ in mind, the relation $S_{\widetilde{h}}(\widetilde{h}_1)>S_{\widetilde{h}}(\widetilde{h}_2)$ gives rise to $S_{\widetilde{h}}(\lambda\widetilde{h}_1)>S_{\widetilde{h}}(\lambda\widetilde{h}_2)$ for any $\lambda>0$, which implies that $\lambda\widetilde{h}_1\succ\lambda\widetilde{h}_2$. This completes the proof. $\Box$
\begin{Proposition}\label{Pr3.3}
(\textit{Monotonicity}). Let $\mu$ be a fuzzy measure on $X$, and $\widetilde{h}_1(x_i)$ and $\widetilde{h}_2(x_i),~(i=1,...,n)$ be two collection of
HOHFEs on $X$. Assume that the HOHFE score function $S_{\widetilde{h}}$ is $\lambda$-invariance, and $(\sigma(1),\sigma(2), . . . ,\sigma(n))$ denotes a permutation of $(1, 2, . . . ,n)$ such that
$\widetilde{h}_1(x_{\sigma(1)})\succ \widetilde{h}_1(x_{\sigma(2)})\succ...\succ \widetilde{h}_1(x_{\sigma(n)})$ and $\widetilde{h}_2(x_{\sigma(1)})\succ \widetilde{h}_2(x_{\sigma(2)})\succ...\succ \widetilde{h}_2(x_{\sigma(n)})$. If $\widetilde{h}_1(x_{\sigma(i)})\succ \widetilde{h}_2(x_{\sigma(i)})$ for $(i=1,...,n)$, then
\begin{eqnarray*}
HOHFEC_\mu(\widetilde{h}_1(x_1),...,\widetilde{h}_1(x_n))\succ HOHFEC_\mu(\widetilde{h}_2(x_1),...,\widetilde{h}_2(x_n)).
\end{eqnarray*}
\end{Proposition}
\textbf{Proof.} The assertion follows by applying Lemma \ref{Le3.2} to Definition \ref{Def3.5}. $\Box$
\begin{Proposition}\label{Pr3.4}
(\textit{Boundedness}). Let $\mu$ be a fuzzy measure on $X$, $\widetilde{h}(x_i),~(i=1,...,n)$ be a collection of
HOHFEs on $X$, and $(\sigma(1),\sigma(2), . . . ,\sigma(n))$ be a permutation of $(1, 2, . . . ,n)$ such that
$\widetilde{h}_1(x_{\sigma(1)})\succ \widetilde{h}_1(x_{\sigma(2)})\succ...\succ \widetilde{h}_1(x_{\sigma(n)})$.
Then, the $\lambda$-invariance property of the HOHFE score function $S_{\widetilde{h}}$ gives rise to
\begin{eqnarray*}
\widetilde{h}_{Max} \succ HOHFEC_\mu(\widetilde{h}_1(x_1),...,\widetilde{h}_1(x_n))\succ \widetilde{h}_{Min},
\end{eqnarray*}
where $\widetilde{h}_{Min}=\min\{\widetilde{h}_1(x_1),...,\widetilde{h}_1(x_n)\}=\widetilde{h}_1(x_{\sigma(n)})$ and $\widetilde{h}_{Max}=\max\{\widetilde{h}_1(x_1),...,\widetilde{h}_1(x_n)\}=\widetilde{h}_1(x_{\sigma(1)})$.
\end{Proposition}
\textbf{Proof.} The application of Proposition \ref{Pr3.3} gives rise to the assertion. $\Box$
\begin{Proposition}\label{Pr3.5}
If for any G-type FS $\widetilde{h_h}$ of HOHFE $\widetilde{h}$, it holds $(\widetilde{h_h}^{(1)}\oplus\widetilde{h_h}^{(2)})\oplus\widetilde{h_h}^{(3)}=\widetilde{h_h}^{(1)}\oplus(\widetilde{h_h}^{(2)}\oplus\widetilde{h_h}^{(3)})$ and $\lambda_1\widetilde{h_h}\oplus\lambda_2\widetilde{h_h}=(\lambda_1+\lambda_2)\widetilde{h_h}$ where $\lambda_1,\lambda_2>0$,
then the HOHF Choquet integral of $\widetilde{h}(x_i),~(i=1,...,n)$ with
respect to $\mu$ introduced in Definition \ref{Def3.5} satisfies
\begin{eqnarray*}
HOHFEC_\mu(\kappa\widetilde{h}_1(x_1)\oplus\widetilde{\overline{h}},...,\kappa\widetilde{h}_1(x_n)\oplus\widetilde{\overline{h}})=
\kappa HOHFEC_\mu(\widetilde{h}_1(x_1),...,\widetilde{h}_1(x_n))\oplus\widetilde{\overline{h}},\quad \kappa>0.
\end{eqnarray*}
\end{Proposition}
\textbf{Proof.} The proof is much like that of Proposition \ref{Pr3.2} and therefore is omitted. $\Box$

\section{MCDM with the HOHF Choquet integral operator}

In this section, we present a new method for MCDM, in which the evaluations of the alternatives are given by HOHFEs and
the interaction among the criteria are allowed.\\
The MCDM procedure with HOHF Choquet integral operator
can be described as follows: suppose that $Y = \{y_1,y_2, . . . ,y_m\}$ is an alternative set of $m$ alternatives. In a MCDM problem, the decision maker is interested in choosing the best one(s) from $Y$ according
to the criteria set $X = \{x_1,x_2, . . . ,x_n\}$. In the HOHFE MCDM, the evaluation of each alternative
on each criterion is a HOHFE. By using the score function $S_{\widetilde{h}}$ introduced in Definition \ref{Def3.2}, all HOHFEs associated with each alternative are re-ordered such that $\widetilde{h}(x_{\sigma(1)})\succ \widetilde{h}(x_{\sigma(2)})\succ...\succ \widetilde{h}(x_{\sigma(n)})$ where $(\sigma(1),\sigma(2), . . . ,\sigma(n))$ is a permutation of $(1, 2, . . . ,n)$. Taking into account the correlations of the HOHFEs, the evaluations of an alternative can be aggregated to its overall evaluation by the HOHF Choquet integral operator. According to a given total order
relation on aggregated HOHFEs, the decision maker can rank those overall evaluations and get the best alternative(s).

\subsection{A practical example}

In what follows, we are going to demonstrate the practicality 
of implementing the proposed concept of HOHFE in a higher order
hesitant fuzzy multi-attribute decision making problem.
\\
{Energy plays an important role in socio-economic development. Thus selecting an appropriate policy for energy
is critical for economic development and environment.  Assume} that there exist five alternatives (energy projects) ${y}_i, (i = 1,2,3,4,5)$ which are invested in accordance with four criteria: $x_1$: technological; $x_2$: environmental; $x_3$: socio-political; $x_4$: economic. 
A number of decision makers are invited to evaluate the performance
of the above five alternatives. For an alternative under {a} criterion, 
the decision makers give their evaluations anonymously in the form of HOHFSs. 
In this regard, the results evaluated by the decision makers are contained in
a higher order hesitant fuzzy decision matrix which is shown in Table 1.

{\scriptsize
\vspace{0cm}

\begin{center}
{{\rm \textbf{Table~1.}}} Higher order hesitant fuzzy decision matrix
\end{center}
\[
\begin{tabular}{|c| c c c c|}\hline
  &  $x_1$    & $x_2$ & $x_3$  & $x_4$
  \\ \hline
${y}_1$ & $\{(0.3, 0.4, 0.5)\}$ & $\{(0.4, 0.5, 0.6), \{0.7, 0.8, 0.9\} \}$ & $\{(0.5, 0.7, 0.7), (0.7, 0.8, 0.9)\}$  & $\{(0.2, 0.3, 0.4), \{0.3, 0.4, 0.5\} \}$  \\
  \hline
${y}_2$ & $\{(0.1, 0.2, 0.3), (0.2, 0.3, 0.4)\}$ & $\{(0.5, 0.6, 0.7), \{0.3, 0.4, 0.5\} \}$ & $\{(0.2, 0.4, 0.6), (0.7, 0.8, 0.9)\}$  & $\{(0.1, 0.4, 0.7), (0.6, 0.7, 0.8)\}$  \\
  \hline
${y}_3$ & $\{(0.1, 0.2, 0.3), \{0.3, 0.4, 0.5\}\}$ & $\{0.7, 0.8, 0.9\}$ & $\{(0.2, 0.3, 0.4), (0.5, 0.6, 0.7)\}$  & $\{(0.4, 0.5, 0.6)\}$  \\
  \hline
${y}_4$ & $\{ 0.2, 0.3, 0.4 \}$ & $\{(0.3, 0.4, 0.5), (0.2, 0.4, 0.6)\}$ & $\{(0.5, 0.6, 0.7), \{0.3, 0.4, 0.5\} \}$  & $\{(0.1, 0.2, 0.3), (0.3, 0.4, 0.5)\}$  \\
  \hline
${y}_5$ & $\{(0.2, 0.4, 0.6), (0.7, 08, 0.9)\}$ & $\{0.7, 0.8, 0.9\}$ & $\{(0.3, 0.4, 0.5)\}$  & $\{(0.5, 0.7, 0.7), (0.7, 0.8, 0.9)\}$  \\
  \hline
 \end{tabular}
\]
\normalsize
\vspace{1cm}

In Table 1, the notation $A = (a_1, a_2 , a_3)$ is used to describe a fuzzy event by the help of {a} triangular fuzzy number in which the values $a_1$,
$a_2$ and $a_3$  denote respectively the smallest possible
value, the most promising value and the largest
possible value, and moreover, $\{\widetilde{h_{\widetilde{h}}}^{(1)},\widetilde{h_{\widetilde{h}}}^{(2)},\widetilde{h_{\widetilde{h}}}^{(3)}\}$ indicates a HFE.
{In this setting and as a consequence, we observe that the second and third elements of the last row of Table 1 are conceptually different. The second array of Table 1 indicates a HFE, while, the third array is used to indicate a triangular fuzzy number.} 

Suppose that the fuzzy measures of criteria $X = \{x_1,x_2,x_3,x_4\}$ are given as follows:
\begin{eqnarray*}
&&\mu(\emptyset)=0;\\
&&\mu(\{x_1\})=0.2,~\mu(\{x_2\})=0.4,~\mu(\{x_3\})=0.3,~\mu(\{x_4\})=0.1;\\
&&\mu(\{x_1,x_2\})=0.4,~\mu(\{x_1,x_3\})=0.2,~\mu(\{x_1,x_4\})=0.1,~\mu(\{x_2,x_3\})=0.3,\\&&\mu(\{x_2,x_4\})=0.3,~\mu(\{x_3,x_4\})=0.1;\\
&&\mu(\{x_1,x_2,x_3\})=0.3,~\mu(\{x_1,x_2,x_4\})=0.1,~\mu(\{x_1,x_3,x_4\})=0.2,~\mu(\{x_2,x_3,x_4\})=0.2;\\
&&\mu({X})=1.
\end{eqnarray*}
For re-ordering HOHFEs given in Table 1, we recall the HOHFE score function $S_{\widetilde{h}}$ introduced in Definition \ref{Def3.2}, where
\begin{eqnarray}\label{leeli0}
S_{\widetilde{h}}(\widetilde{h})=\frac{1}{|\widetilde{h}|}\sum_{\widetilde{h_{h}}\in \widetilde{h}}S_{\widetilde{{h_{h}}}}(\widetilde{{h_{h}}}).
\end{eqnarray}
Here, the G-type FS score function $S_{\widetilde{{h_{h}}}}$ is taken as 
\begin{itemize}
\item  (For triangular fuzzy number):
\begin{eqnarray}\label{leeli}
S_{\widetilde{{h_{h}}}}({(a_1, a_2 , a_3)})=\frac{1}{3}(a_1+a_2+a_3),
\end{eqnarray}
which was proposed by Lee and Li in \cite{leeli};
\item  (For hesitant fuzzy element):
\begin{eqnarray}\label{hfe-scor}
S_{\widetilde{{h_{h}}}}(\{\widetilde{h_{\widetilde{h}}}^{(1)},...,\widetilde{h_{\widetilde{h}}}^{(|\widetilde{h}|)}\})=\frac{1}{|\widetilde{h}|}(\{\widetilde{h_{\widetilde{h}}}^{(1)}+...+\widetilde{h_{\widetilde{h}}}^{(|\widetilde{h}|)}\}),
\end{eqnarray}
\end{itemize}
Although, the corresponding score functions of HFEs and triangular fuzzy numbers may differ from each other, but in this example, both of them have a same rule.
However, by using the HOHFE score function $S_{\widetilde{h}}$ given by (\ref{leeli0}), we re-arrange the HOHFEs corresponding to each criterion in
descending order as follows:
\begin{eqnarray*}
&&y_1:\quad \widetilde{h}(x_{\sigma(1)}=x_3)\succ\widetilde{h}(x_{\sigma(2)}=x_2)\succ\widetilde{h}(x_{\sigma(3)}=x_1)\succ\widetilde{h}(x_{\sigma(4)}=x_4);\\
&&y_2:\quad \widetilde{h}(x_{\sigma(1)}=x_3)\succ\widetilde{h}(x_{\sigma(2)}=x_4)\succ\widetilde{h}(x_{\sigma(3)}=x_2)\succ\widetilde{h}(x_{\sigma(4)}=x_1);\\
&&y_3:\quad \widetilde{h}(x_{\sigma(1)}=x_2)\succ\widetilde{h}(x_{\sigma(2)}=x_4)\succ\widetilde{h}(x_{\sigma(3)}=x_3)\succ\widetilde{h}(x_{\sigma(4)}=x_1);\\
&&y_4:\quad \widetilde{h}(x_{\sigma(1)}=x_3)\succ\widetilde{h}(x_{\sigma(2)}=x_2)\succ\widetilde{h}(x_{\sigma(3)}=x_1)\succ\widetilde{h}(x_{\sigma(4)}=x_4);\\
&&y_5:\quad \widetilde{h}(x_{\sigma(1)}=x_2)\succ\widetilde{h}(x_{\sigma(2)}=x_4)\succ\widetilde{h}(x_{\sigma(3)}=x_1)\succ\widetilde{h}(x_{\sigma(4)}=x_3).
\end{eqnarray*}
If we employ the HOHF Choquet integral operator, we then are able to aggregate all HOHFEs in the i-$th$ row of
the HOHFE decision matrix into an overall value. For instance,
\begin{eqnarray*}\nonumber
&&\widetilde{h}_{y_1}=HOHFEC_\mu(\widetilde{h}(x_1),...,\widetilde{h}(x_4))=\\
&& [(\mu(A_{\sigma(1)})-\mu(A_{\sigma(0)}))\widetilde{h}(x_{\sigma(1)})]\oplus[(\mu(A_{\sigma(2)})-\mu(A_{\sigma(1)}))\widetilde{h}(x_{\sigma(2)})]\\
&& \oplus[(\mu(A_{\sigma(3)})-\mu(A_{\sigma(2)}))\widetilde{h}(x_{\sigma(3)})]\oplus[(\mu(A_{\sigma(4)})-\mu(A_{\sigma(3)}))\widetilde{h}(x_{\sigma(4)})]\\
&&=[(\mu(\{x_{\sigma(1)}\})-\mu(\emptyset))\widetilde{h}(x_{\sigma(1)})]\oplus[(\mu(\{x_{\sigma(1)},x_{\sigma(2)}\})-\mu(\{x_{\sigma(1)}\}))
\widetilde{h}(x_{\sigma(2)})]\\
&& \oplus[(\mu(\{x_{\sigma(1)},x_{\sigma(2)},x_{\sigma(3)}\})-\mu(\{x_{\sigma(1)},x_{\sigma(2)}\}))\widetilde{h}(x_{\sigma(3)})]
\oplus[(\mu(X)-\mu(\{x_{\sigma(1)},x_{\sigma(2)},x_{\sigma(3)}\}))\widetilde{h}(x_{\sigma(4)})]\\
&&=[(\mu(\{x_{3}\})-\mu(\emptyset))\widetilde{h}(x_{3})]\oplus[(\mu(\{x_{3},x_{2}\})-\mu(\{x_{3}\}))
\widetilde{h}(x_{2})]\\
&& \oplus[(\mu(\{x_{3},x_{2},x_{1}\})-\mu(\{x_{3},x_{2}\}))\widetilde{h}(x_{1})]
\oplus[(\mu(X)-\mu(\{x_{3},x_{2},x_{1}\}))\widetilde{h}(x_{4})]\\
&&=[(0.3-0)\{(0.5, 0.7, 0.7), (0.7, 0.8, 0.9)\}]\oplus[(0.3-0.3)
\{(0.4, 0.5, 0.6), \{0.7, 0.8, 0.9\} \}]\\
&& \oplus[(0.3-0.3)\{(0.3, 0.4, 0.5)\}]
\oplus[(1-0.3)\{(0.2, 0.3, 0.4), \{0.3, 0.4, 0.5\} \}].
\end{eqnarray*}
Let us now apply the following arithmetic operations 
\begin{itemize}
\item  (For triangular fuzzy number \cite{kaops}):
\begin{eqnarray*}
&&\lambda A = \lambda (a_1, a_2 , a_3)=(\lambda a_1, \lambda a_2 , \lambda a_3), \\
&&A \oplus B = (a_1, a_2 , a_3)\oplus (b_1, b_2 , b_3)=(a_1+b_1-a_1b_1, a_2+b_2-a_2b_2 , a_3+b_3-a_3b_3);
\end{eqnarray*}
\item  (For hesitant fuzzy element {\cite{far-book}}):
\begin{eqnarray*}
&&\lambda  \{\widetilde{h_{\widetilde{h}}}^{(1)},...,\widetilde{h_{\widetilde{h}}}^{(|\widetilde{h}|)}\}=
  \{\lambda \widetilde{h_{\widetilde{h}}}^{(1)},...,\lambda \widetilde{h_{\widetilde{h}}}^{(|\widetilde{h}|)}\},\quad \lambda\geq0,\\
&& \{\widetilde{h_{\widetilde{h_1}}}^{(1)},...,\widetilde{h_{\widetilde{h_1}}}^{(|\widetilde{h}|)}\}\oplus  
\{\widetilde{h_{\widetilde{h_2}}}^{(1)},...,\widetilde{h_{\widetilde{h_2}}}^{(|\widetilde{h}|)}\}
\\
&&\quad\quad=
  \{ \widetilde{h_{\widetilde{h_1}}}^{(1)}+\widetilde{h_{\widetilde{h_2}}}^{(1)}-
  \widetilde{h_{\widetilde{h_1}}}^{(1)}\widetilde{h_{\widetilde{h_2}}}^{(1)}
  ,..., \widetilde{h_{\widetilde{h_1}}}^{(|\widetilde{h}|)}+\widetilde{h_{\widetilde{h_2}}}^{(|\widetilde{h}|)}
  -\widetilde{h_{\widetilde{h_1}}}^{(|\widetilde{h}|)}\widetilde{h_{\widetilde{h_2}}}^{(|\widetilde{h}|)}
  \},
\end{eqnarray*}
\end{itemize}
 to the latter findings. Then, one achieves that
\begin{eqnarray*}\nonumber
\widetilde{h}_{y_1}
&=&[\{(0.15, 0.21, 0.21), (0.21, 0.24, 0.27)\}]\oplus
[\{(0.14, 0.21, 0.28), \{0.21, 0.28, 0.35\} \}]
\\
&=&
\{(0.27,    0.37 ,   0.43),(0.32,    0.40 ,   0.47), \{0.21, 0.28, 0.35\}\}.
\end{eqnarray*}
The HOHFE score value of $\widetilde{h}_{y_1}$ is then equal to
\begin{eqnarray*}
S_{\widetilde{h}}(\widetilde{h}_{y_1})=\frac{1}{|\widetilde{h}_{y_1}|}\sum_{\widetilde{h_{{h}_{y_1}}}\in \widetilde{h}_{y_1}}S_{\widetilde{{h_{h}}}}(\widetilde{{h_{{h}_{y_1}}}})=0.3456.
\end{eqnarray*}
Similarly, the other HOHFE score values are obtained as:
\begin{eqnarray*}
S_{\widetilde{h}}(\widetilde{h}_{y_2})=0.4915,\quad S_{\widetilde{h}}(\widetilde{h}_{y_3})=0.4364,\quad
S_{\widetilde{h}}(\widetilde{h}_{y_4})=0.2739,\quad S_{\widetilde{h}}(\widetilde{h}_{y_5})=0.5601.
\end{eqnarray*}
Therefore,
\begin{eqnarray*}
S_{\widetilde{h}}(\widetilde{h}_{y_5})> S_{\widetilde{h}}(\widetilde{h}_{y_2})>
S_{\widetilde{h}}(\widetilde{h}_{y_3})> S_{\widetilde{h}}(\widetilde{h}_{y_1})>S_{\widetilde{h}}(\widetilde{h}_{y_4}),
\end{eqnarray*}
which are led to 
\begin{eqnarray*}
y_5\succ y_2 \succ y_3 \succ y_1 \succ y_4,
\end{eqnarray*}
and hence the most appropriate energy project is $y_5$.

\subsection{Comparisons and further discussions}

Here, we adopt a multiple criteria group decision making problem from \cite{peng1} which is originally based on extending TODIM method by encountering the Choquet integral within a multiset
hesitant fuzzy environment.\\
Peng et al. \cite{peng1} considered an investment company for investing in a project. They supposed five alternatives  ${y}_1$:  car company; ${y}_2$: food company; ${y}_3$: computer company; ${y}_4$: arms company; and ${y}_5$: TV company which need to be invested. In this setting, the desired decision has to be made according to the four criteria including $x_1$: the environment impact 
which is referred to as the impact on the companys environment 
$x_2$: the risk factor  
including product risk
and development environment risk; 
$x_3$: the growth prospects
including increased profitability and returns; and $x_4$: social-political impact that is referred to as the governments and local
residents support for company. 
\\
Peng et al. \cite{peng1} {used multiset hesitant fuzzy sets to evaluate alternatives by two decision makers.} In this contribution to have a complete picture about the performance of the proposed approach and that of Peng et al. \cite{peng1}, we re-state Peng et al's \cite{peng1} decision matrix in the following form of higher order hesitant fuzzy decision matrix: 


{\scriptsize
\vspace{0cm}

\begin{center}
{{\rm \textbf{Table~2.}}} Higher order hesitant fuzzy decision matrix
\end{center}
\[
\begin{tabular}{|c| c c c c|}\hline
  &  $x_1$    & $x_2$ & $x_3$  & $x_4$
  \\ \hline
${y}_1$ & $\{(0.4, 0.5, 0.7)\}$ & $\{(0.5, 0.5, 0.8)\}$ & $\{(0.6, 0.6, 0.9)\}$  & $\{0.5, 0.6\}$  \\
  \hline
${y}_2$ & $\{(0.6, 0.7, 0.8)\}$ & $\{0.5, 0.6\}$ & $\{(0.6, 0.7, 0.7)\}$  & $\{0.4, 0.5\}$  \\
  \hline
${y}_3$ & $\{0.6, 0.8\}$ & $\{(0.2, 0.3, 0.5)\}$ & $\{0.6, 0.6\}$  & $\{0.5, 0.7\}$  \\
  \hline
${y}_4$ & $\{(0.5, 0.5, 0.7)\}$ & $\{0.4, 0.5\}$ & $\{0.8, 0.9\}$  & $\{(0.3, 0.4, 0.5)\}$  \\
  \hline
${y}_5$ & $\{0.6, 0.7\}$ & $\{0.5, 0.7\}$ & $\{0.7, 0.8\}$  & $\{(0.3, 0.3, 0.4)\}$  \\
  \hline
 \end{tabular}
\]
\normalsize
\vspace{1cm}
\\
some of its arrays are in form of triangular fuzzy numbers and some of them are HFEs. Needless to say that preserving both forms of arrays 
appears to maintain the implication of HOHFEs.


According to  Peng et al's \cite{peng1} assumption, the fuzzy measures of criteria $X = \{x_1,x_2,x_3,x_4\}$ are as follows:
\begin{eqnarray*}
&&\mu(\emptyset)=0;\\
&&\mu(\{x_1\})=0.40,~\mu(\{x_2\})=0.25,~\mu(\{x_3\})=0.37,~\mu(\{x_4\})=0.20;\\
&&\mu(\{x_1,x_2\})=0.60,~\mu(\{x_1,x_3\})=0.70,~\mu(\{x_1,x_4\})=0.56,~\mu(\{x_2,x_3\})=0.68,\\&&\quad \quad\quad \mu(\{x_2,x_4\})=0.43,~\mu(\{x_3,x_4\})=0.54;\\
&&\mu(\{x_1,x_2,x_3\})=0.88,~\mu(\{x_1,x_2,x_4\})=0.75,~\mu(\{x_1,x_3,x_4\})=0.84,~\mu(\{x_2,x_3,x_4\})=0.73;\\
&&\mu(X)=1.
\end{eqnarray*}
We re-arrange the HOHFEs corresponding to each criterion in
descending order by the use of HOHFE score function $S_{\widetilde{h}}$ given by (\ref{leeli0}) as follows:
\begin{eqnarray*}
&&y_1:\quad \widetilde{h}(x_{\sigma(1)}=x_3)\succ\widetilde{h}(x_{\sigma(2)}=x_2)\succ\widetilde{h}(x_{\sigma(3)}=x_1)\succ\widetilde{h}(x_{\sigma(4)}=x_4);\\
&&y_2:\quad \widetilde{h}(x_{\sigma(1)}=x_1)\succ\widetilde{h}(x_{\sigma(2)}=x_3)\succ\widetilde{h}(x_{\sigma(3)}=x_2)\succ\widetilde{h}(x_{\sigma(4)}=x_4);\\
&&y_3:\quad \widetilde{h}(x_{\sigma(1)}=x_1)\succ\widetilde{h}(x_{\sigma(2)}=x_3)\succ\widetilde{h}(x_{\sigma(3)}=x_4)\succ\widetilde{h}(x_{\sigma(4)}=x_2);\\
&&y_4:\quad \widetilde{h}(x_{\sigma(1)}=x_3)\succ\widetilde{h}(x_{\sigma(2)}=x_1)\succ\widetilde{h}(x_{\sigma(3)}=x_2)\succ\widetilde{h}(x_{\sigma(4)}=x_4);\\
&&y_5:\quad \widetilde{h}(x_{\sigma(1)}=x_3)\succ\widetilde{h}(x_{\sigma(2)}=x_1)\succ\widetilde{h}(x_{\sigma(3)}=x_2)\succ\widetilde{h}(x_{\sigma(4)}=x_4).
\end{eqnarray*}
Now, by the help of HOHF Choquet integral operator, we aggregate all HOHFEs in the i-$th$ row of
the HOHFE decision matrix into an overall value, for example,
\begin{eqnarray*}\nonumber
&&\widetilde{h}_{y_1}=HOHFEC_\mu(\widetilde{h}(x_1),...,\widetilde{h}(x_4))=\\
&& [(\mu(A_{\sigma(1)})-\mu(A_{\sigma(0)}))\widetilde{h}(x_{\sigma(1)})]\oplus[(\mu(A_{\sigma(2)})-\mu(A_{\sigma(1)}))\widetilde{h}(x_{\sigma(2)})]\\
&& \oplus[(\mu(A_{\sigma(3)})-\mu(A_{\sigma(2)}))\widetilde{h}(x_{\sigma(3)})]\oplus[(\mu(A_{\sigma(4)})-\mu(A_{\sigma(3)}))\widetilde{h}(x_{\sigma(4)})]\\
&&=[(\mu(\{x_{\sigma(1)}\})-\mu(\emptyset))\widetilde{h}(x_{\sigma(1)})]\oplus[(\mu(\{x_{\sigma(1)},x_{\sigma(2)}\})-\mu(\{x_{\sigma(1)}\}))
\widetilde{h}(x_{\sigma(2)})]\\
&& \oplus[(\mu(\{x_{\sigma(1)},x_{\sigma(2)},x_{\sigma(3)}\})-\mu(\{x_{\sigma(1)},x_{\sigma(2)}\}))\widetilde{h}(x_{\sigma(3)})]
\oplus[(\mu(X)-\mu(\{x_{\sigma(1)},x_{\sigma(2)},x_{\sigma(3)}\}))\widetilde{h}(x_{\sigma(4)})]\\
&&=[(\mu(\{x_{3}\})-\mu(\emptyset))\widetilde{h}(x_{3})]\oplus[(\mu(\{x_{3},x_{2}\})-\mu(\{x_{3}\}))
\widetilde{h}(x_{2})]\\
&& \oplus[(\mu(\{x_{3},x_{2},x_{1}\})-\mu(\{x_{3},x_{2}\}))\widetilde{h}(x_{1})]
\oplus[(\mu(X)-\mu(\{x_{3},x_{2},x_{1}\}))\widetilde{h}(x_{4})]\\
&&=\{(0.0765,    0.0795,    0.1205), \{0.0100 ,   0.0120\}\}.
\end{eqnarray*}
By applying the same procedure, we achieve 
\begin{eqnarray*}
&&\widetilde{h}_{y_2}=\{(0.0550,    0.0642,    0.0672), \{0.0533,    0.0645\} \},\\ 
&& \widetilde{h}_{y_3}=\{(0.0110,    0.0165,    0.0275), \{0.0650,    0.0750\} \},\\ 
&& \widetilde{h}_{y_4}=\{(0.0210,    0.0230,    0.0310), \{0.0713,    0.0830\} \},\\ 
&& \widetilde{h}_{y_5}=\{(0.0060,    0.0060,    0.0080), \{0.1330,    0.1633\} \}.
\end{eqnarray*}
Moreover, the HOHFE score function (\ref{leeli}) results in 
\begin{eqnarray*}
S_{\widetilde{h}}(\widetilde{h}_{y_1})=0.2985,\quad S_{\widetilde{h}}(\widetilde{h}_{y_2})=0.3041,\quad
S_{\widetilde{h}}(\widetilde{h}_{y_3})=0.1950,\quad S_{\widetilde{h}}(\widetilde{h}_{y_4})=0.2293,\quad 
S_{\widetilde{h}}(\widetilde{h}_{y_5})=0.3163.
\end{eqnarray*}
Therefore,
\begin{eqnarray*}
S_{\widetilde{h}}(\widetilde{h}_{y_5})> S_{\widetilde{h}}(\widetilde{h}_{y_2})>
S_{\widetilde{h}}(\widetilde{h}_{y_1})> S_{\widetilde{h}}(\widetilde{h}_{y_4})>S_{\widetilde{h}}(\widetilde{h}_{y_3}).
\end{eqnarray*}
These findings indicate that
\begin{eqnarray*}
y_5\succ y_2 \succ y_1 \succ y_4 \succ y_3,
\end{eqnarray*}
and hence the most appropriate energy project is $y_5$.

In what follows, we are going to verify the validity of proposed technique compared to the other existing techniques including 
Xu's \cite{comp20},
Wei's \cite{comp22}, Zhang et al.'s \cite{comp24}, Chen et al.'s \cite{comp27}, Xu's \cite{comp28},
Farhadinia's \cite{comp30}, Zhang and Wei's \cite{comp31}, Zhang and Xu's \cite{comp32}, Wang et al.'s \cite{comp36} and Peng et al.'s \cite{peng1} from a perspective of performance. Indeed, from the performance stand point of view, 
we intend to discuss the resulted rankings of techniques macroscopy, and not microscopy, that is, we do not intend to investigate the pros and cons of each technique.
\\
{Before going more in depth with future analysis, let us first 
re-state here from Farhadinia \cite{sort} the algorithm being implemented to sort decision making techniques based on their outcomes.}

{
\begin{Algorithm}\label{alg} \cite{sort} (\textit{Sorting of decision making techniques based on their outcomes})
\begin{description}
  \item[Step 1.] Take into account each outcome of decision making techniques (i.e., the resulted ranking order of alternatives) as an individual preference 
  $R_{k}$ $(k=1,...,m)$
   together with 
   the associated dominance-vector 
   $\pi_{R_{k}}$.
\item[Step 2.] Add up all the individual
preference matrices component-wisely to get the collective preference matrix $\overline{R}=[\overline{r}_{ij}]_{n\times n}$. 
\item[Step 3.] Extract the collective preference from the matrix $\overline{R}=[\overline{r}_{ij}]_{n\times n}$ with respect to the collective majority decision rule.
\item[Step 4.] Characterize the corresponding \textit{dominance-vector hesitant fuzzy sets} to the individual and the collective preferences. Then, determine the importance weight of decision making techniques in accordance with the distance value of their individual preference from the collective preference using a corresponding distance measure.
\end{description}
\end{Algorithm}
}


Now, for determining the importance weights
of decision making techniques according to their distance values of their individual rankings from the collective one, we {implement here {Algorithm \ref{alg} above}}. {Indeed, Algorithm \ref{alg}} is allocated to 
assess the best outcome(s) without regarding of how
the preference ordering is obtained or from what numerical
values one concludes the preference ordering.
\\
Now, by considering the total ranking
order $R$ on $Y = \{y_1, y_2, y_3,y_4,y_5\}$, we can correspond to each of them the dominance-vectors $\pi_{R_k}$ (for $k=X1,W1,...,Pro$) which are given in Table {3}.

{\small
\vspace{0cm}

\begin{center}
{{\rm \textbf{Table~{3}.}}}  The dominance-vectors corresponding to the different techniques.
\end{center}
\[
\begin{tabular}{|c| c|c|}\hline
 Techniques  & Ranking~ orders & Dominance-vector
  \\ \hline
$(e_{X1})$: Xu's \cite{comp20}~ technique & $R_{X1}$: $y_4\succ y_5 \succ y_2 \succ y_1 \succ y_3$ & $\pi_{R_{X1}}=(2,3,1,4,5)$ \\
  \hline
$(e_{W1})$: Wei's \cite{comp22}~ technique & $R_{W1}$: $y_2\succ y_5 \succ y_4 \succ y_1 \succ y_3$ & $\pi_{R_{W1}}=(2,5,1,3,4)$ \\
  \hline
$(e_{Z1})$: Zhang et al.'s \cite{comp24}~ technique & $R_{Z1}$: $y_5\succ y_2 \succ y_1 \succ y_4 \succ y_3$ & $\pi_{R_{Z1}}=(3,4,1,2,5)$ \\
  \hline
$(e_{C})$: Chen et al.'s \cite{comp27}~ technique & $R_{C}$: $y_5\succ y_1 \succ y_2 \succ y_4 \succ y_3$ & $\pi_{R_{C}}=(4,3,1,2,5)$ \\
  \hline
$(e_{X2})$: Xu's \cite{comp28}~ technique & $R_{X2}$: $y_5\succ y_2 \succ y_4 \succ y_1 \succ y_3$ & $\pi_{R_{X2}}=(2,4,1,3,5)$ \\
  \hline
$(e_{F})$: Farhadinia's \cite{comp30}~ technique  & $R_{F}$: $y_5\succ y_1 \succ y_2 \succ y_4 \succ y_3$ & $\pi_{R_{F}}=(4,3,1,2,5)$ \\
  \hline  
$(e_{Z2})$: Zhang and Wei's \cite{comp31}~ technique & $R_{Z2}$: $y_5\succ y_1 \succ y_2 \succ y_4 \succ y_3$ & $\pi_{R_{Z2}}=(4,3,1,2,5)$ \\
  \hline 
$(e_{Z3})$: Zhang and Xu's \cite{comp32}~ technique  & $R_{Z3}$: $y_5\succ y_1 \succ y_2 \succ y_3 \succ y_4$ & $\pi_{R_{Z3}}=(4,3,2,1,5)$ \\
  \hline 
$(e_{W2})$: Wang et al.'s \cite{comp36}~  technique & $R_{W2}$: $y_5\succ y_1 \succ y_2 \succ y_3 \succ y_4$ & $\pi_{R_{W2}}=(4,3,2,1,5)$ \\
  \hline 
$(e_{P})$: Peng et al.'s \cite{peng1}~ technique & $R_{P}$: $y_5\succ y_2 \succ y_1 \succ y_3 \succ y_4$ & $\pi_{R_{P}}=(3,4,2,1,5)$\\
  \hline \hline
$(e_{Pro})$: The~ proposed~ technique & $R_{Pro}$: $y_5\succ y_2 \succ y_1 \succ y_4 \succ y_3$ & $\pi_{R_{Pro}}=(3,4,1,2,5)$ \\
  \hline 
 \end{tabular}
\]
\normalsize
\vspace{1cm}

By considering all the individual preference matrices
$R_{k}$ $(k =X1,W1,...,Pro)$, {for instance, $R_{X1}$: $y_4\succ y_5 \succ y_2 \succ y_1 \succ y_3$} which is given by 
\begin{eqnarray*}
R_{X1}=\left(
          \begin{array}{ccccc}
            0 & -1 & 1 & -1 & -1\\
            1 & 0 & 1 & -1 & -1 \\
            -1& -1 & 0 & -1 &-1 \\
            1 & 1 & 1 &0 &1 \\
            1 & 1 & 1 &-1 &0 \\
          \end{array}
        \right),
\end{eqnarray*}
the collective preference $\overline{R}$ is then achieved
through the aggregation process in which 
all the individual preference matrices are added up
component-wisely.
The result is as follows:
\begin{eqnarray*}
\overline{R}=\left(
          \begin{array}{ccccc}
      0 &   -1 &   11&     5&   -11\\
      1&     0&    11 &    9&    -9\\
    -11 &  -11 &    0 &   -5&  -11\\
     -5&   -9 &    5&     0&    -9\\
     11&     9  &  11&     9&     0\\
          \end{array}
        \right),
\end{eqnarray*}
which returns 
$y_5\succ y_2 \succ y_1 \succ y_4 \succ y_3$ as the  collective preference.
\\
Now, with the distance 
\begin{eqnarray}\label{sp1}
&&{\mathcal{D}}({\pi_{R_k}},{\pi_{\overline{R}}})=\max\{ \max_{\pi_{R_k}}\min_{\pi_{\overline{R}}}
|\pi_{R_k},\pi_{\overline{R}}|~,~
 \min_{\pi_{R_k}}\max_{\pi_{\overline{R}}}
|\pi_{R_k},\pi_{\overline{R}}|\}, \quad (k=X1,W1,...,Pro)
\end{eqnarray}
at hand, we may obtain the distance of individual preferences from the collective preference as those given in Table {4}.

\newpage
{\small
\vspace{4 cm}

\begin{center}
{{\rm \textbf{Table~{4}.}}}  The distance of individual preferences from the collective preference.
\end{center}
\[
\begin{tabular}{|c| c|c|}\hline
 Techniques  & Ranking~ orders & 
 Distance values
  \\ \hline \hline
$(e_{X1})$: Xu's \cite{comp20}~ technique & $R_{X1}$: $y_4\succ y_5 \succ y_2 \succ y_1 \succ y_3$ & ${{\cal{D}}}(\pi_{R_{X1}},{\pi}_{\overline{R}})=4$ \\
  \hline
$(e_{W1})$: Wei's \cite{comp22}~ technique & $R_{W1}$: $y_2\succ y_5 \succ y_4 \succ y_1 \succ y_3$ & ${{\cal{D}}}(\pi_{R_{W1}},{\pi}_{\overline{R}})=4$\\
  \hline
$(e_{Z1})$: Zhang et al.'s \cite{comp24}~ technique & $R_{Z1}$: $y_5\succ y_2 \succ y_1 \succ y_4 \succ y_3$ & ${{\cal{D}}}(\pi_{R_{Z1}},{\pi}_{\overline{R}})=0$ \\
  \hline
$(e_{C})$: Chen et al.'s \cite{comp27}~ technique & $R_{C}$: $y_5\succ y_1 \succ y_2 \succ y_4 \succ y_3$ & ${{\cal{D}}}(\pi_{R_{C}},{\pi}_{\overline{R}})=2$ \\
  \hline
$(e_{X2})$: Xu's \cite{comp28}~ technique & $R_{X2}$: $y_5\succ y_2 \succ y_4 \succ y_1 \succ y_3$ & ${{\cal{D}}}(\pi_{R_{X2}},{\pi}_{\overline{R}})=2$ \\
  \hline
$(e_{F})$: Farhadinia's \cite{comp30}~ technique  & $R_{F}$: $y_5\succ y_1 \succ y_2 \succ y_4 \succ y_3$ & ${{\cal{D}}}(\pi_{R_{F}},{\pi}_{\overline{R}})=2$ \\
  \hline  
$(e_{Z2})$: Zhang and Wei's \cite{comp31}~ technique & $R_{Z2}$: $y_5\succ y_1 \succ y_2 \succ y_4 \succ y_3$ &${{\cal{D}}}(\pi_{R_{Z2}},{\pi}_{\overline{R}})=2$\\
  \hline 
$(e_{Z3})$: Zhang and Xu's \cite{comp32}~ technique  & $R_{Z3}$: $y_5\succ y_1 \succ y_2 \succ y_3 \succ y_4$ & ${{\cal{D}}}(\pi_{R_{Z3}},{\pi}_{\overline{R}})=4$ \\
  \hline 
$(e_{W2})$: Wang et al.'s \cite{comp36}~  technique & $R_{W2}$: $y_5\succ y_1 \succ y_2 \succ y_3 \succ y_4$ & ${{\cal{D}}}(\pi_{R_{W2}},{\pi}_{\overline{R}})=4$ \\
  \hline 
$(e_{P})$: Peng et al.'s \cite{peng1}~ technique & $R_{P}$: $y_5\succ y_2 \succ y_1 \succ y_3 \succ y_4$ & ${{\cal{D}}}(\pi_{R_{P}},{\pi}_{\overline{R}})=2$\\
  \hline \hline
$(e_{Pro})$: The~ proposed~ technique & $R_{Pro}$: $y_5\succ y_2 \succ y_1 \succ y_4 \succ y_3$ & ${{\cal{D}}}(\pi_{R_{Pro}},{\pi}_{\overline{R}})=0$ \\
  \hline 
 \end{tabular}
\]
\normalsize
\vspace{1cm}

{The larger} distance ${{\cal{D}}}$ indicates the more
dis-similar between each individual preference and the collective preference. This implies that the larger value of distance {shows
lower performance 
 of} decision-making technique.
In view of Table {4}, we are able to sort the considered techniques which are labelled by $k = X1,W1,..., Pro$ as the following:
\begin{eqnarray*}
\{ (e_{Pro}), (e_{Z1})\}>\{ (e_{C}),(e_{X2}), (e_{F}), (e_{Z2}), (e_{P})\}>\{ (e_{X1}), (e_{W1}), (e_{Z3}), (e_{W2})\}.
\end{eqnarray*}
The above results verify that among the above decision-making techniques, the proposed technique can be chosen as the one of the most suitable
techniques.

{However, these} findings indicate that although the procedure of obtaining the most appropriate alternative using different techniques may be
different, but their outcomes can be used {for constructing an admissible assessment which was called here as the collective preference.}

\section{Conclusion}

The current contribution proposed the higher order hesitant fuzzy (HOHF) Choquet integral operator that not only considers the importance of the elements, but also it can reflect the correlations among the elements. Then, we employed the HOHF Choquet integral operator to aggregate HOHFEs in a higher order hesitant fuzzy MCDM problem.
Eventually, a comparative analysis of different techniques with the HOHF Choquet integral-based technique presented, and it indicates that the latter technique is enough efficient.   
In future work, we plan to study the other aggregation
operators in the higher order hesitant fuzzy setting for handling
MCDM with higher order hesitant fuzzy information. 
In addition, since the application potentials
of HOHFS are diverse, it can be investigated in other domains such as clustering, pattern recognition, image processing, etc.



\end{document}